\documentclass[10pt,twocolumn,letterpaper]{article}
\usepackage[pagenumbers]{cvpr} 

\definecolor{cvprblue}{rgb}{0.21,0.49,0.74}
\usepackage[pagebackref,breaklinks,colorlinks,allcolors=cvprblue]{hyperref}
\usepackage{multirow}    
\usepackage{amssymb}     
\usepackage{booktabs}    
\usepackage{graphicx}    
\usepackage{array}       
\usepackage{makecell}

\def\confName{CVPR}
\def\confYear{2026}

\title{RS-Prune: Training-Free Data Pruning at High Ratios for Efficient Remote Sensing Diffusion Foundation Models}

\author{
Fan Wei$^1$ \quad 
Runmin Dong$^{2}$\thanks{Corresponding authors.} \quad 
Yushan Lai$^4$ \quad 
Yixiang Yang$^6$ \quad 
Zhaoyang Luo$^4$ \\
Jinxiao Zhang$^1$ \quad 
Miao Yang$^1$ \quad 
Shuai Yuan$^5$ \quad 
Jiyao Zhao$^3$ \quad 
Bin Luo$^4$ \quad 
Haohuan Fu$^{1,3,4}$\footnotemark[1]
\\[1ex]
$^1$Department of Earth System Science, Tsinghua University \\
$^2$School of Artificial Intelligence, Sun Yat-sen University \\
$^3$National Supercomputing Center in Shenzhen \\
$^4$Tsinghua Shenzhen International Graduate School, Tsinghua University \\
$^5$Department of Geography, The University of Hong Kong \\
$^6$Wangxuan Institute of Computer Technology, Peking University
}

\begin{document}
\maketitle
\begin{abstract}

Diffusion-based remote sensing (RS) generative foundation models are essential for various downstream tasks, such as super-resolution and image reconstruction. However, these models rely on large amounts of globally representative data, which often contain redundancy, noise, and class imbalance, reducing training efficiency and sometimes preventing convergence.
Existing RS diffusion foundation models typically aggregate multiple classification datasets or apply simplistic deduplication, overlooking the distributional requirements of generation modeling and the inherent heterogeneity of RS imagery. 

To address these limitations, we propose a training-free, two-stage data pruning approach that quickly selects a high-quality subset under high pruning ratios, enabling a preliminary foundation model to converge rapidly and serve as a versatile backbone for generation, downstream fine-tuning, and other applications. Our method jointly considers local information content with global scene-level diversity and representativeness. First, an entropy-based criterion efficiently removes low-information samples. Next, leveraging RS scene classification datasets as reference benchmarks, we perform scene-aware clustering with stratified sampling to improve clustering effectiveness while reducing computational costs on large-scale unlabeled data. Finally, by balancing cluster-level uniformity and sample representativeness, the method enables fine-grained selection under high pruning ratios while preserving overall diversity and representativeness. 
Experiments on both curated RS datasets and large-scale global imagery show that, even after pruning 85\% of the training data, our method significantly improves convergence and generation quality.
Furthermore, diffusion foundation models trained with our method consistently achieve state-of-the-art performance across downstream tasks, including super-resolution and semantic image synthesis. This data pruning paradigm offers practical guidance and empirical insight for the development of RS generative foundation models. 

\end{abstract}

\begin{figure}[t]
    \centering
    \vspace{-5pt}
    \includegraphics[width=0.99\linewidth]{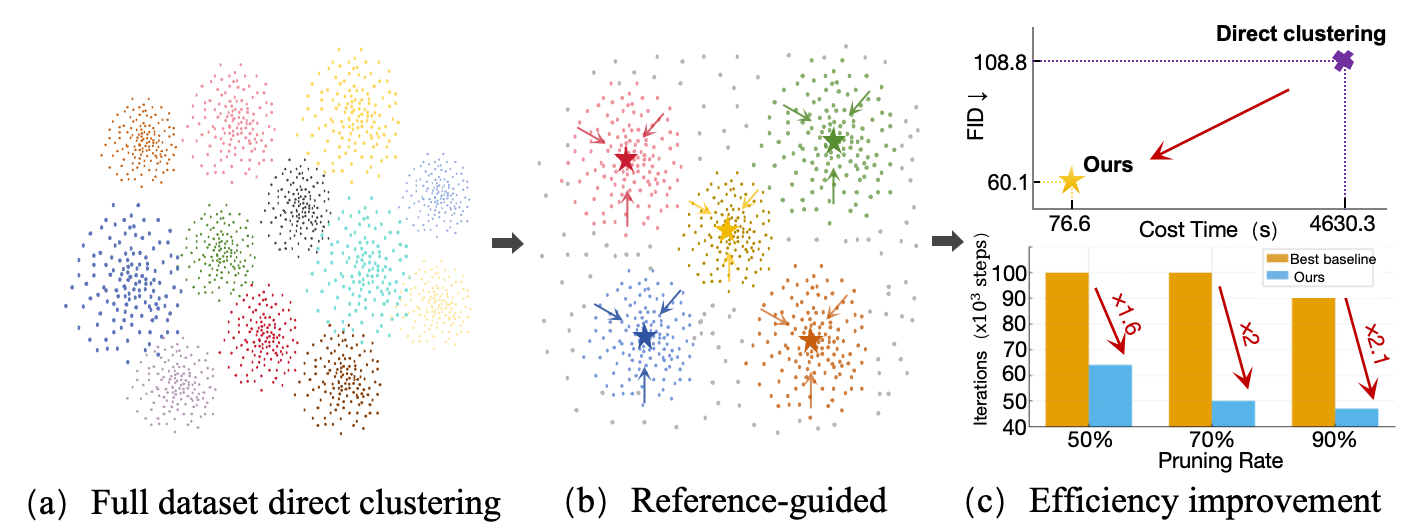}
    \caption{Illustration of our reference‐guided clustering and efficiency gains. (a) Direct clustering on the full unlabeled dataset is computationally expensive. (b) Our reference‐guided strategy selects a representative, scene-aware subset around reference centroids. (c) Compared with full-dataset clustering, our method achieves much lower FID at substantially reduced cost time. And
     compared with best baseline, ours requires 1.9× to 2.1× fewer training iterations across different pruning ratios.}
    \label{fig:head}
    \vspace{-10pt}
\end{figure}    
\section{Introduction}
\label{sec:intro}

In recent years, generation models, especially diffusion models~\cite{Peebles2022DiT, Ho2020DDPM}, have achieved remarkable progress in fields such as computer vision~\cite{Richard2021Binaural}, medical imaging~\cite{Song2021ScoreMedical}, and remote sensing (RS)~\cite{Dong2024BridgeSR}. Within the RS domain, generation models have been widely applied to data augmentation, image reconstruction, super-resolution, and high-resolution image synthesis, supporting practical applications in urban planning, land-use monitoring, and disaster response~\cite{Borana2023UrbanSustainability}. A powerful RS 
diffusion foundation model can provide a robust data and modeling backbone to further enhance these applications.

However, the effective training of RS diffusion foundation models critically depends on the quality and distribution of training data. The emergence of large-scale open-source datasets (e.g., Git-10M~\cite{Liu2025Text2Earth}, RS5M~\cite{Zhang2024RS5M}) provides valuable resources, yet it also introduces several critical challenges, including image redundancy, low-quality samples (e.g., noise and cloud cover), class imbalance~\cite{Cheng2017NWPU}, and scene homogeneity~\cite{Xia2017AID}. These issues not only hinder the training efficiency of RS foundation models but also limit the effectiveness of the resulting pretrained models on downstream tasks.

Recently, several studies in RS generation tasks have made preliminary explorations into data processing. For instance, 
RSDiff employs size cropping and noise augmentation~\cite{Sebaq2024RSDiff}.
WHU-RS19~\cite{Balestra2025WHURS19} removes cloud and open-ocean regions to reduce low information content. Super-resolution works~\cite{Huang2015Urban100} often prioritize urban areas. Nevertheless, these methods mainly rely on rule-based or simplistic strategies to eliminate redundant data, without considering the specific dependence of generation models on data distribution~\cite{Goodfellow2014GAN} or the inherent characteristics of RS imagery, such as diversity, heterogeneity, and class balance. 
Moreover, many data pruning methods developed in the computer vision domain rely on scoring mechanisms~\cite{Yang2024CLIPDataSelection} from supervised pretrained models~\cite{Pleiss2020AUCMargin} or retrieval of nearby images based on ImageNet benchmarks~\cite{Oquab2023DINOv2}. In contrast, a standardized labeled dataset for RS, comparable to ImageNet~\cite{Deng2009ImageNet} in computer vision and spanning multiple resolutions and modalities, is currently lacking. Overall, systematic research on data selection for RS diffusion foundation models remains scarce.
%


\begin{figure*}
  \centering
    \includegraphics[width=\linewidth]{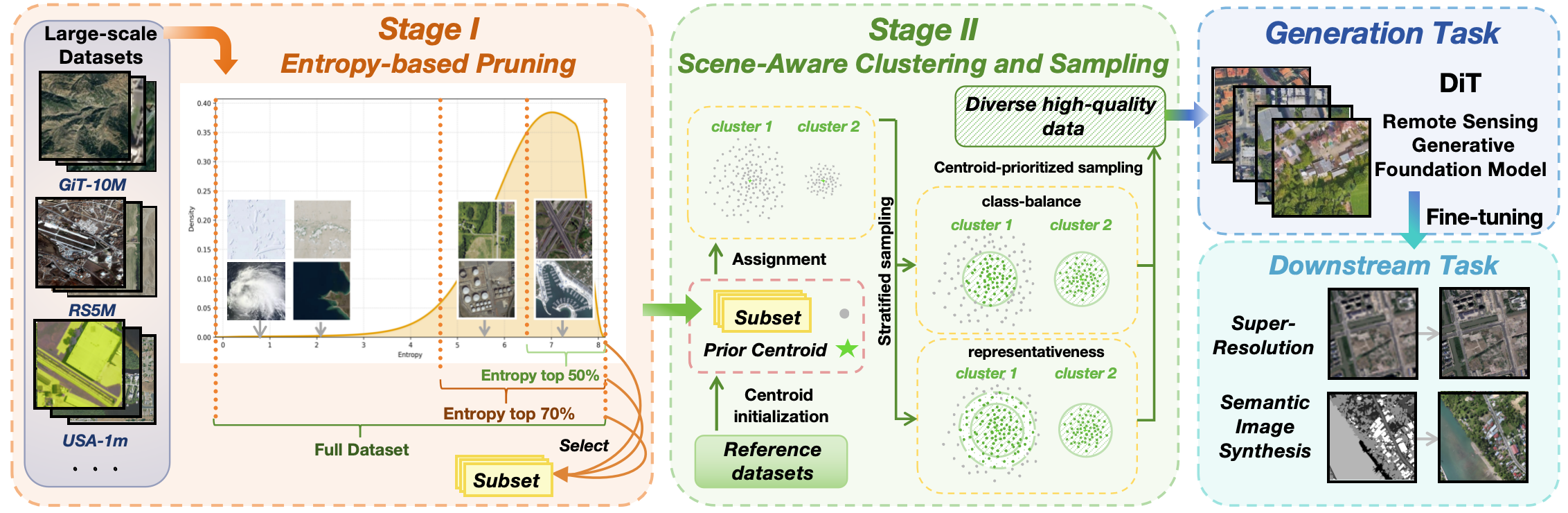}
    \caption{Overview of our multi-stage data pruning method for Remote Sensing generative foundation models.}
    \label{fig:overview}
  \hfill
  \vspace{-8pt}
\end{figure*}

The effectiveness of diffusion foundation models depends more on the quality and distribution of the training data than on dataset size alone. Recent studies~\cite{Briq2024DataPruning} have shown that diffusion models retain strong generation performance even when a large fraction (e.g., 90\% of ImageNet samples) of the training data is removed, highlighting that a significant portion of the data contributes minimally. In the RS domain, raw data often contains substantial redundancy and noise, meaning that merely increasing the dataset size does not guarantee proportional performance gains.
If generative models are trained on large volumes of data without appropriate selection, low-quality and redundant samples can not only slow convergence and increase computational and time costs, but also introduce distribution shifts that ultimately degrade model performance.

To address these issues, we propose an efficient, trainning-free data pruning approach for RS generative diffusion models, which can construct a diverse subset under high pruning ratios, enabling diffusion models to converge significantly faster with better generative quality. 

More specifically, targeting redundancy, low quality samples, and class imbalance, we systematically explore data pruning strategies across both global-scale scenarios (GiT-10M, RS5M) and urban-scale settings (our constructed USA-1m multispectral dataset). The method proceeds along two complementary dimensions:
1) Information dimension. We employ entropy-based pruning to rapidly discard low texture or homogeneous large-area regions, thereby reducing redundancy and compressing subsequent computation costs. 2) Diversity and representativeness dimension. We introduce scene–aware clustering with stratified sampling. Using existing RS scene-classified datasets as reference, we perform over-clustering on the reference datasets to obtain hundreds of cluster centroids, effectively avoiding the high computational cost of full-scale clustering and full dataset training. Large-scale unlabeled datasets are then assigned to these predefined clusters in the clustering space. Samples are subsequently selected using a combination of class-balanced allocation and centroid-prioritized sampling, which preferentially chooses samples near cluster centers to preserve scene-representative characteristics while maintaining diversity.

Extensive experiments reveal three key observations: First, both datasets specifically constructed for RS generation models and globally collected datasets contain substantial redundancy, and appropriate data pruning accelerates convergence while yielding models that outperform those trained on the full datasets. Second, entropy-based pruning consistently removes low quality and highly homogeneous samples, providing stable improvements across varying pruning rates. Third, optimal pruned subsets are obtained by combining entropy-based filtering with strategies that preserve diversity and representativeness at the scene level, ensuring that informative and representative samples are retained for effective model training.


Our main contributions in this paper can be summarized as follows:

\begin{itemize}[leftmargin=*, label=\textbullet]
\item We systematically explore data pruning for RS diffusion foundation models and propose a training-free, two-stage selection strategy that considers data heterogeneity, diversity, and representativeness, enabling faster convergence and improved generation performance under high pruning ratios.

\item We introduce a reference-guided clustering method, performing pre-clustering on curated scene-classified datasets to preserve the diversity of scene cluster centers while avoiding the computational cost of clustering massive unlabeled datasets and full-dataset training.

\item Extensive experiments demonstrate that our method can rapidly construct a high-quality training subset with substantial scale reduction. Training on this subset not only accelerates model convergence significantly but also outperforms prior state-of-the-art methods in pre-training generation quality while achieving gains on downstream tasks.
%

\end{itemize}

\section{Related Work}
\label{sec:formatting}


\subsection{Diffusion foundation Models}


Diffusion foundation models~\cite{Saharia2022Imagen} play a pivotal role in image generation, data synthesis, and image reconstruction. For instance, Stable Diffusion (SD) 1.5~\cite{Rombach2022LDM} leverages latent-space representations together with a UNet backbone, serving as a lightweight generation foundation model that provides strong pretrained initialization for ControlNet~\cite{Zhang2023ControlNet} and related low-level vision tasks~\cite{Saharia2022Palette}. More recently, Transformer-based architectures such as DiT~\cite{Peebles2022DiT} have emerged, while models like SD3~\cite{Esser2024RectifiedFlow} and Flux~\cite{BFL2024FluxRepo} offer substantially stronger generation capabilities and higher fidelity, further advancing the scalability and versatility of diffusion frameworks.


In the RS domain, diffusion foundation models have also started to gain attention. To address the unique properties of RS imagery, such as multi-spectral and multi-resolution observations, geospatial information, and global coverage, many works have explored pre-training or fine-tuning foundation models on RS datasets~\cite{Tang2024CRSDiff,Toker2024SatSynth,Xiao2023EDiffSR}. For example, DiffusionSat~\cite{Khanna2023DiffusionSat} introduced geolocation as conditioning information and fine-tuned Stable Diffusion 1.5 on multi-source RS data, supporting multiple RS generation tasks. Meanwhile, 
models such as SR3~\cite{Saharia2021SR3} and CDM~\cite{Ho2021CDM}, demonstrated strong super-resolution performance on natural images, providing a foundation for their adaptation to RS.
Despite these advances, existing RS generation models typically aggregate multiple classification datasets or apply simplistic deduplication~\cite{Liu2024DiffusionSurvey}, thereby overlooking the distributional requirements of generation foundation modeling as well as the inherent heterogeneity and diversity of RS imagery. Consequently, the field lacks systematic exploration of data pruning strategies specifically designed to address the characteristics of remote sensing data and the requirements of RS generative foundation models.




\subsection{Data Pruning Methods}

Training data are critical to constructing RS diffusion foundation models. However, existing studies~\cite{Liu2023RemoteCLIP, Sebaq2024RSDiff} in this field have adopted simplistic deduplication and preprocessing strategies, such as removing cloud and ocean regions~\cite{Balestra2025WHURS19} and prioritizing urban areas~\cite{Zhang2024A2MAE}. These approaches are insufficient to address the sensitivity of generation models to data distribution, heterogeneity, and class imbalance.

Current data pruning methods can be broadly categorized into three types. First, data-valuation methods assign an importance score to each sample and select samples accordingly. For example, MoSo~\cite{Tan2023MoSo} estimates the change in empirical risk when a sample is removed. Although such methods are generally efficient, their performance can be affected by group effects and may lack generalization in complex real-world settings. Second, distribution-based methods rely on the geometric structure of the dataset. For instance, Moderate-DS~\cite{moderate} selected samples near the median. CCS~\cite{Zheng2022CCS} balanced data distribution and sample importance during selection. Finally, optimization-based methods leverage optimization techniques to guide sample pruning, such as temporal dual-depth scoring~\cite{Zhang2024TDDS}, gradient matching~\cite{Killamsetty2021GradMatch}, scalable self-supervised pruning metrics~\cite{Sorscher2022DataPruning}, influence functions~\cite{KohLiang2017Influence}, and bilevel optimization~\cite{Borsos2020CoresetBilevel}.

Most of these approaches are designed for supervised datasets and rely on scores generated by pre-trained supervised models. In practice, they typically require training or computation over the full dataset, which requires labeled data and is computationally expensive, making them difficult to scale to large, mostly unlabeled RS datasets. 

Beyond explicit pruning algorithms, recent foundation models have also explored large-scale data pruning pipelines. For example, DINOv2~\cite{Oquab2023DINOv2} constructs its curated LVD-142M dataset by performing PCA-hash and copy-detection based deduplication, followed by self-supervised retrieval based on ImageNet benchmarks. Similarly, V-JEPA 2~\cite{assran2025vjepa2selfsupervisedvideo} adopts a retrieval and reweighting strategy to align large-scale video data with target distributions.

However, such data curation pipelines typically rely on target datasets or distributions and require large-scale clustering or retrieval over the entire datasets. In remote sensing, there is no benchmark similar to ImageNet that adequately captures the diversity of global RS imagery, and  RS diffusion foundation models are particularly sensitive to the diversity and representativeness of semantic categories. To bridge this gap, we systematically explore training-free, unsupervised data pruning strategies tailored to the characteristics of RS imagery and the requirements of RS diffusion foundation models.
\section{METHOD}

\subsection{Workflow}
RS generative foundation models depend on large-scale, globally collected datasets that provide high quality, diversity, and representativeness. However, existing RS datasets often lack these characteristics, leading to slower model convergence, suboptimal generation performance, and insufficient capability to support various low-level downstream tasks. To address this, we propose a two-stage data pruning method (\cref{fig:overview}) guided by two key principles:

\textbf{1) Informational value} Cloud-covered or excessively homogeneous RS images are inevitable in global data collection. Such images are typically low-quality, contain limited informative content, and exhibit substantial redundancy. Images with higher information content, capturing meaningful structures and fine details, are prioritized for selection. This process effectively removes low-information and trivially homogeneous scenes, such as vast desert expanses, thereby preserving heterogeneity in the selected subset.

\textbf{2) Scene diversity and centroid representativeness}
While maintaining the overall semantic distribution, rare scenes are preferentially preserved. For scenes with abundant samples, candidates are ranked by their similarity to the cluster centroid, and the nearest-centroid samples are selected. This strategy not only ensures high quality and de-redundancy but also aligns with expert priors reflected in reference datasets, producing a subset that is both distributionally representative and diverse.

\subsection{Stage I: Entropy-based Pruning}

We measure the information value of each image by computing its global Shannon entropy~\cite{Shannon1948}. 
Specifically, for an image $I$, we compute its grayscale entropy $H(I)$ to capture the diversity of pixel intensities:
{
\begin{equation}
H(I) = - \sum_{k=0}^{L-1} p_k \log p_k,
\end{equation}
}
where $p_k$ denotes the empirical probability of intensity level $k$ among $L$ possible levels. 
Images with $H(I) < \tau$ are discarded, as they typically correspond to invalid regions (e.g., sensor noise) or low-variation scenes (e.g., clouds, open ocean, deserts, or saturated exposures). 
This pruning step substantially reduces dataset size while preserving high-information candidate samples.

\subsection{Stage II: Scene-Aware Clustering and Sampling}
The remote sensing domain lacks a universally adopted, comprehensive benchmark comparable to ImageNet in the natural image domain. To address this, we leverage multiple expert-curated RS classification datasets as a composable bank of clustering priors, covering various scene types such as urban areas, cropland, water bodies, forests, and transportation infrastructure. Unlike approaches that apply label-free clustering and sampling directly on a massive generic corpus, we first establish stable scene centroids on this prior bank. Importantly, instead of computing centroids separately for each labeled category, we conduct over-clustering across the entire prior dataset to obtain diverse and representative centroids. Subsequently, for the large-scale unlabeled RS dataset, samples are aligned to these centroids and selected according to centroid-prioritized sampling, ensuring both representativeness and diversity. This pipeline uses expert-curated classification datasets to derive more representative cluster centroids, thereby enhancing diversity and representativeness in the selected subset, while simultaneously reducing the computational cost of clustering the full unlabeled large-scale dataset.

\paragraph{Prior Centroid Construction.}
We employ a unified feature extractor $f(\cdot)$, namely Git-RSCLIP~\cite{Liu2025Text2Earth}, with $\ell_{2}$ normalization to embed all images from standard datasets. Git-RSCLIP is specifically pretrained on large-scale remote sensing corpora and achieves substantially better performance than previous RS-oriented CLIP variants such as RemoteCLIP and GeoRSCLIP.

{
 \begin{equation}
 \mathbf{z}_x=\frac{f(x)}{\lVert f(x)\rVert_2}\quad\text{for each }x\in\mathcal{D}_{\text{ref}},
 \end{equation}
}

where $\mathcal{D}_{\text{ref}}$ is the collection of reference (prior) datasets. $x$ is an image sample of the datasets.  
$f(\cdot)$ is a pretrained image encoder.
$\mathbf{z}_x$ is the $\ell_2$-normalized embedding of $x$. We then perform $K$-means clustering on the unit hypersphere to identify representative scene centroids. Let 
$\mathcal{M}$ denote the set of learned centroids:
{
\begin{equation}
\mathcal{M} \;=\; \{\boldsymbol{\mu}_k\}_{k=1}^{K}, \quad \lVert \boldsymbol{\mu}_k \rVert_2 = 1,
\end{equation}
}
where $K$ is the number of clusters and $\boldsymbol{\mu}_k$ is the $k$-th centroid in the feature space. 

\paragraph{Cluster Assignment and Candidate Pooling.}
For each unlabeled sample $x \in \mathcal{D}_{\mathrm{u}}$, we compute its cosine similarity to all prior centroids:

{
\begin{equation}
s_k(x) \;=\; \langle f(x),\, \boldsymbol{\mu}_k \rangle,
\qquad k=1,\dots,K,
\end{equation}
}
where $s_k(x)$ is the similarity score. $f(x)$ denotes the feature embedding of $x$. Each sample is assigned to the cluster with the highest similarity:
{
\begin{equation}
\hat{z}(x) \;=\; \arg\max_{k} \; s_k(x),
\end{equation}
}
where $\hat{z}(x)$ is the hard cluster label. The sample is then added into the candidate pool corresponding to its assigned cluster, denoted $P_k $. This assignment procedure scales linearly with both the number of unlabeled samples and the number of centroids, making it efficient for large-scale datasets.

\paragraph{Cluster-Aware Stratified Sampling.}
Given a total sampling budget $B$, we combine class-balanced allocation with centroid-prioritized sampling.
\begin{itemize}[leftmargin=1.2em,labelsep=0.5em]
  \item \textbf{Class-balanced allocation.}
 Let $\{P_k\}_{k=1}^{K}$ denote the candidate pools for $K$ clusters, and let $q=\left\lfloor B/K \right\rfloor$.
We assign each cluster a quota of $q$ samples to ensure coverage across clusters. Rare clusters with fewer than $q$ samples retain all candidates, preserving diversity.

  \item \textbf{Centroid-prioritized sampling.}
Within each cluster $k$, we rank samples $x \in P_k$ by their similarity $s_k(x)$ to the cluster centroid $\boldsymbol{\mu}_k$ and select the top-$q$ samples:
{
\begin{equation}
S_k \;=\; \operatorname*{Top\text{-}q}_{x \in P_k}\; s_k(x).
\end{equation}
}
If $|P_k|<q$, we set $S_k=P_k$ and reallocate the remaining budget by selecting additional samples from the global remainder:
$P_{\mathrm{rem}}=\bigcup_{j=1}^{K} \big(P_j \setminus S_j\big)$
in descending order of similarity until the total number of selected samples equals $B$, i.e., $B=\sum_{k=1}^{K} |S_k|$.
\end{itemize}


\subsection{Complexity Analysis}
In stage I, the overall complexity is approximately $\mathcal{O}(N)$, as each image is evaluated individually for information value.

In stage II, we have:
1) Building clustering priors on a small standardized reference set is done once, so its cost is negligible relative to the full pipeline.
2) Assigning each image to the most similar centroid requires computing similarities to $K$ centroids in a $f_d$-dimensional feature space, giving $\mathcal{O}(N K f_d)$.
3) Within each cluster, ranking images by similarity and sampling $q$ samples requires sorting, which has a complexity of $\mathcal{O}\!\left(\frac{N}{K} \log \frac{N}{K}\right)$ per cluster, resulting in $\mathcal{O}\!\left(N \log \frac{N}{K}\right)$.
Since $K$ and $f_d$ are small constants in practice, the end-to-end complexity can be considered $\mathcal{O}(N\log N)$.
The method does not involve training deep models. The core operations are vector arithmetic, similarity computation, and sorting.
Consequently, the pipeline is highly scalable and practically deployable for data pruning over large-scale RS datasets.

\section{EXPERIMENTS}

\subsection{Datasets and evaluation}
\begin{figure*}[t]
  \centering
  \begin{subfigure}[t]{0.33\linewidth}
    \centering
    \includegraphics[width=\linewidth]{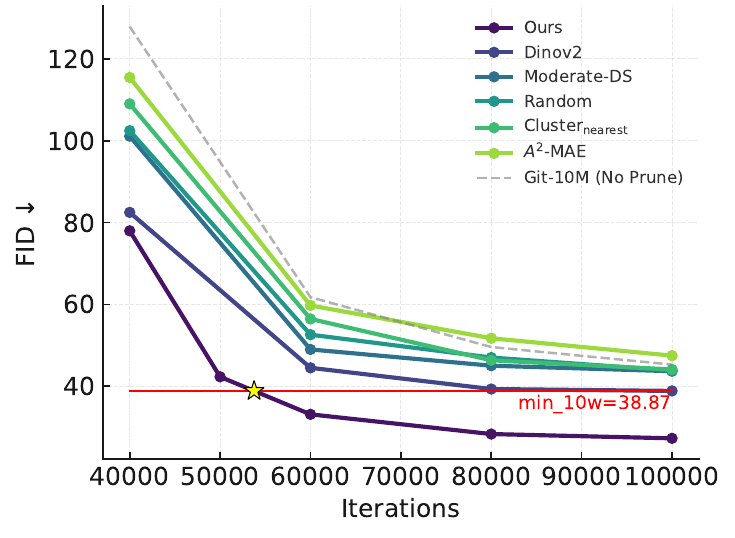}
    \caption{Pruning rate 50\%}
    \label{fig4-1}
  \end{subfigure}\hfill
  \begin{subfigure}[t]{0.33\linewidth}
    \centering
    \includegraphics[width=\linewidth]{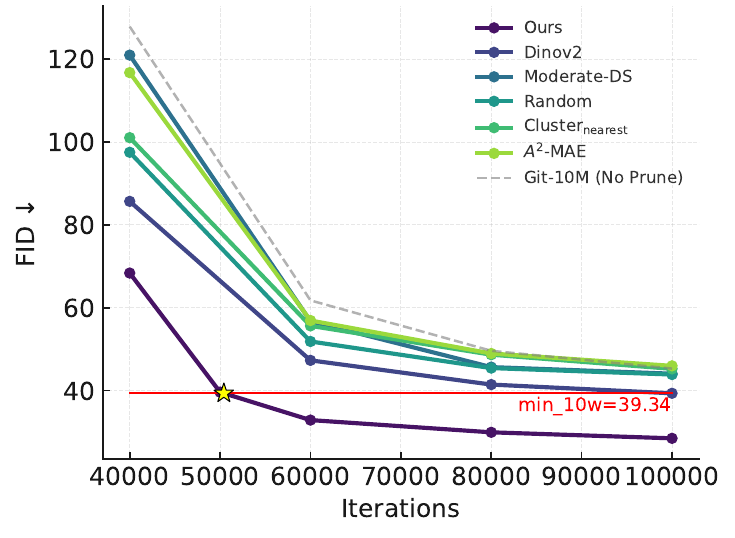}
    \caption{Pruning rate 70\%}
    \label{fig4-2}
  \end{subfigure}\hfill
  \begin{subfigure}[t]{0.33\linewidth}
    \centering
    \includegraphics[width=\linewidth]{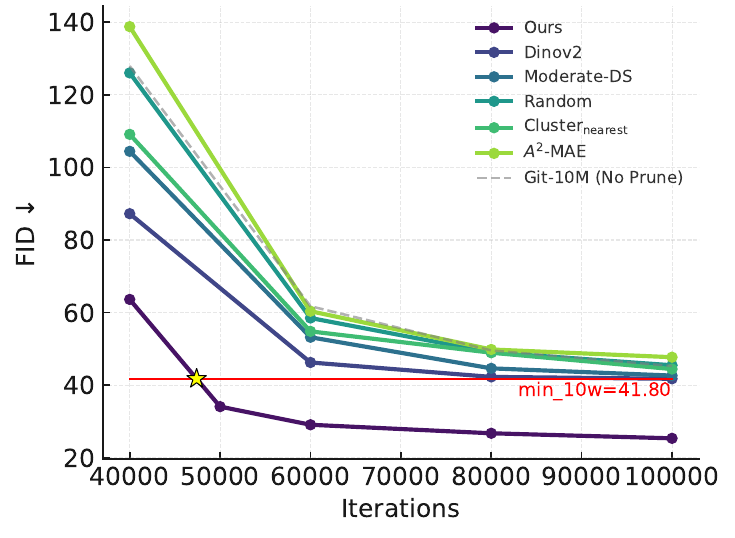}
    \caption{Pruning rate 90\%}
    \label{fig4-3}
  \end{subfigure}
  \caption{Comparison of generation performance (FID) across different pruning ratios on Git-10M.}
  \label{fig-robustness-to-corruption}
\end{figure*}

\begin{table*}[t]
\centering
\caption{Experimental results with different data pruning methods on Git-10M and USA-1m. We report results at a pruning ratio of 70\%.}
\label{tab:gen-metrics-two-datasets}
\resizebox{1\textwidth}{!}{%
\renewcommand{\arraystretch}{1}
\small
\begin{tabular}{c|ccccc|ccccc}
\toprule
\multirow{3}{*}{\makecell{Method}} &
\multicolumn{5}{c|}{\textbf{Git-10M}} &
\multicolumn{5}{c}{\textbf{USA-1m}} \\
\cline{2-11}
& \multicolumn{1}{c}{Generation} & \multicolumn{2}{c}{SR} & \multicolumn{2}{c|}{SIS} &
  \multicolumn{1}{c}{Generation} & \multicolumn{2}{c}{SR} & \multicolumn{2}{c}{SIS} \\
\cline{2-11}
& FID$\downarrow$
& FID$\downarrow$ & LPIPS$\downarrow$
& FID$\downarrow$ & LPIPS$\downarrow$
& FID$\downarrow$
& FID$\downarrow$ & LPIPS$\downarrow$
& FID$\downarrow$ & LPIPS$\downarrow$ \\
\midrule
Full Dataset           & 45.30 & 89.25 & 0.3912 & 125.79 & 0.6027 & 242.52 & 125.90 & 0.4909 & 327.33 & 0.8212 \\
Random                & 43.87 & 89.25 & 0.3912 & 127.00 & 0.6200 & 255.11 & 126.63 & 0.4827 & 236.68 & 0.6698 \\
Moderate\mbox{-}DS~\cite{moderate}    & 44.07 & 92.40 & 0.3952 & 132.85 & 0.6222 & 199.24 & 124.86 & 0.5098 & 233.13 & 0.6730 \\
$\mathrm{Cluster}_{\text{nearest}}$~\cite{Briq2024DataPruning}
                      & 45.24 & 89.58 & 0.3898 & 133.18 & 0.6237 & 201.36 & 150.11 & 0.5870 & 256.40 & 0.6804 \\
Dinov2~\cite{Oquab2023DINOv2}                & 39.34 & 91.76 & 0.3930 & 137.55 & 0.6121 & 180.98 & 127.73 & 0.5472 & 238.80 & 0.6761 \\
A$^{2}$\mbox{-}MAE~\cite{Zhang2024A2MAE}    & 45.96 & 90.39 & 0.3938 & 130.64 & 0.6013 & 360.74 & 148.36 & 0.6028 & 256.39 & 0.6993 \\
\cline{1-11}
\textbf{Ours}         & \textbf{28.46} & \textbf{87.98} & \textbf{0.3893} & \textbf{122.08} & \textbf{0.5967} & \textbf{175.93} & \textbf{122.00} & \textbf{0.4779} & \textbf{199.82} & \textbf{0.6682} \\
\bottomrule
\end{tabular}
} 
\vspace{-8pt}
\end{table*}

To evaluate the effectiveness of our data pruning approach for RS generation tasks, we conduct experiments on both global-scale and urban-scale datasets. Specifically, we use two representative global-scale optical datasets (i.e., GIT-10M and RS5M) that were carefully curated for RS generative foundation models, as well as a large-scale urban-scale multispectral dataset (i.e., USA-1m) collected in this work based on U.S. urban boundary products~\cite{Li2020GUB}. The global datasets provide broad coverage and diversity, while the urban dataset offers finer spatial resolution and richer texture information. This design allows us to assess the effectiveness and generalizability of our data pruning strategy across datasets with varying resolutions, modalities, volumes, and sensors.





\begin{itemize}[leftmargin=1.2em,labelsep=0.5em]
  \item \textbf{Git-10M}~\cite{Liu2025Text2Earth} A \emph{global-scale} dataset comprising 10.5M satellite images (RGB) with a spatial resolution of 0.5 to 128\,m, which spans multiple continents and geographic regions, covering diverse land-cover types such as urban areas, croplands, forests, mountains, and deserts. 

  \item \textbf{RS5M}~\cite{Zhang2024RS5M} A \emph{global-scale} dataset initially collected from 11 publicly available image–text pair datasets, containing RS images (RGB) from different regions, resolutions, and scene types. Due to large variations in image size and relatively low data quality, we apply simple preprocessing and filtering, resulting in a curated subset of about 1.04M images.

  \item \textbf{USA-1m (multispectral)} Our self-constructed, urban-scale remote sensing dataset consisting of 8.77M four-channel, high-resolution ($1$\,m) images, derived from the USDA National Agriculture Imagery Program (NAIP)~\cite{Robinson2019LargeScale}. The dataset covers major urban areas across 45 U.S. states, with a total footprint of $1.34339\times10^{6}$\,km$^{2}$.
  
\end{itemize}

For each of the aforementioned datasets, we selected 5,000 high-quality images for subsequent metric computation for the generation task.



In Stage II, we adopt five scene-classification datasets with distinct characteristics: NWPU-RESISC45~\cite{Cheng2017NWPU}, UC Merced Land-Use~\cite{Yang2010UCMerced}, AID~\cite{Xia2017AID}, WHU-RS19~\cite{Balestra2025WHURS19}, and RSD46-WHU~\cite{Long2017RSD46}. Detailed information regarding their resolution, dataset size, number of classes, and data sources can be found in supplementary materials.


To comprehensively assess the impact of training data on RS generation models, we evaluate their generation capability directly on the test set and further use downstream-task metrics to evaluate the capability of the pretrained models, i.e., super-resolution (SR) and semantic image synthesis (SIS). The evaluation metrics include FID~\cite{Heusel2017FID} and LPIPS~\cite{Zhang2018LPIPS}. Lower LPIPS and FID values indicate superior generation performance. 
%


\begin{itemize}[leftmargin=1.2em,labelsep=0.5em]

\item \textbf{SECOND-SR}
We construct a super-resolution dataset based on the high-resolution SECOND semantic change detection benchmark~\cite{Yang2020SECOND}. Low-resolution (LR) images are generated by first downsampling and then upsampling the original high-resolution (HR) images. The dataset contains 4,662 paired LR–HR images, each with a spatial size of $512{\times}512$ pixels, covering six major land-cover classes.

\item \textbf{OpenEarthMap}~\cite{Xia2023OpenEarthMap}
 A global benchmark designed for semantic segmentation. It consists of 5,000 RS images across six continents at 0.25 to 0.5\,m resolutions. Labels are annotated with eight land-cover categories.

\end{itemize}

\subsection{Implementation Details}
All models are implemented using the official DiT framework~\cite{Peebles2022DiT}, which is a transformer-based architecture using its github repository\footnote{https://github.com/facebookresearch/DiT}. 
We adopt the DiT-XL/2 backbone for both generation pretraining and downstream tasks. All experiments are conducted with $256{\times}256$ input size, a global batch size of 256, and AdamW optimizer with a fixed learning rate of $1\times10^{-4}$.

Training is distributed across 4 NVIDIA H100 GPUs.
\textbf{Base VAE} For Git\mbox{-}10M and RS5M, we use the publicly released Stable Diffusion VAE (\texttt{sd-vae-ft-ema})\footnote{https://huggingface.co/stabilityai/sd-vae-ft-ema}. For USA\mbox{-}1M, we train a custom four-channel VAE from scratch and select the checkpoint with the lowest validation loss.
\textbf{Generation pretraining} DiT is trained in an unconditional setting. For each dataset, we run $40$K to $100$K diffusion steps and models are trained from scratch. \textbf{Downstream fine-tuning} For super-resolution and semantic image synthesis, 
we initialize from the pretrained DiT checkpoints and fine\mbox{-}tune for 5,000 steps per task.

\begin{table*}[tb!]
\centering
\caption{Results of scene-aware clustering and sampling at high pruning ratios. The subset refers to the Stage I output used as input for Stage II.}
\setlength{\tabcolsep}{3mm}
\resizebox{0.9\textwidth}{!}{
\renewcommand{\arraystretch}{1.1}
\begin{tabular}{c|>{\small}c|c|c|ccc|ccc}
\toprule
\multirow{2}{*}{\makecell{\\Pruning\\ratio}} &
\multirow{2}{*}{\makecell{\\\normalsize Subset}} &
\multirow{2}{*}{\makecell{\\Stage\\I}} &
\multirow{2}{*}{\makecell{\\Stage\\II}} &
\multicolumn{3}{c|}{\textbf{Git-10M}} &
\multicolumn{3}{c}{\textbf{RS5M}} \\
\cmidrule(lr){5-7}\cmidrule(lr){8-10}
& & & &
\multicolumn{1}{c}{generation} & \multicolumn{2}{c|}{SIS} &
\multicolumn{1}{c}{generation} & \multicolumn{2}{c}{SIS} \\
& & & &
FID$\downarrow$ & FID$\downarrow$ & LPIPS$\downarrow$ &
FID$\downarrow$ & FID$\downarrow$ & LPIPS$\downarrow$ \\
\midrule
\multirow{4}{*}{85\%}
& $\mathcal{D}^{H\uparrow}_{15\%}$ & \checkmark &  &
68.7814 & 125.7938 & 0.6044 &
59.4955 & 126.7798 & 0.5966 \\
& $\mathcal{D}^{H\uparrow}_{30\%}$ & \checkmark & \checkmark &
\textbf{61.3269} & \textbf{120.5167} & \textbf{0.5982} &
\textbf{58.9813} & \textbf{115.6430} & \textbf{0.5902} \\
& $\mathcal{D}^{H\uparrow}_{50\%}$ & \checkmark & \checkmark &
64.4233 & 139.9440 & 0.6139 &
73.7312 & 132.6490 & 0.6014 \\
& $\mathcal{D}^{H\uparrow}_{70\%}$ & \checkmark & \checkmark &
72.3580 & 130.6310 & 0.6139 &
75.4898 & 121.8777 & 0.5941 \\
\midrule
\multirow{4}{*}{90\%}
& $\mathcal{D}^{H\uparrow}_{10\%}$ & \checkmark &        &
70.9027 & 121.9285 & 0.6043 &
65.4067 & 124.6616 & 0.6034 \\
& $\mathcal{D}^{H\uparrow}_{30\%}$ & \checkmark & \checkmark &
63.6400 & 123.0029 & 0.6106 &
64.2536 & 124.8206 & 0.6004 \\
& $\mathcal{D}^{H\uparrow}_{50\%}$ & \checkmark & \checkmark &
64.2870 & 144.4404 & 0.6173 &
71.3744 & 130.5425 & 0.6094 \\
& $\mathcal{D}^{H\uparrow}_{70\%}$ & \checkmark & \checkmark &
77.2366 & 131.4739 & 0.6094 &
79.8913 & 128.3239 & 0.5970 \\
\bottomrule
\end{tabular}
}
\label{tab:scene}
\vspace{-3pt}
\end{table*}

\subsection{Comparison Results}

We compare the proposed method with three groups of pruning strategies: basic pruning methods (e.g., Random, Moderate-DS~\cite{moderate}), data pruning methods for foundation models (e.g., Dinov2~\cite{Oquab2023DINOv2}, A$^{2}$\mbox{-}MAE~\cite{Zhang2024A2MAE}), and a recent approach specifically developed for generation models ($\mathrm{Cluster}_{\text{nearest}}$~\cite{Briq2024DataPruning}). To examine their effectiveness across varying scene complexities, experiments are conducted on a global-scale dataset (Git-10M) and an urban-scale dataset (USA-1m). To ensure comparability, all models are trained for an equivalent number of iterations, 100K for Git-10M and 40K for USA-1m, and the best results during training are reported. Beyond generation evaluation on the test set, pretrained models are further assessed on two downstream tasks, i.e., SR and SIS, with the top fine-tuning results recorded.

In subsequent experiments, we observe that the evaluation metrics on the generative task exhibit consistent trends with the evaluation metrics on different downstream tasks. This suggests that generative metrics can effectively characterize the capability of the pretrained models. Therefore, in some experiments, we use the generative metrics as a representative metrics for assessing pretrained model capacity.

As illustrated in \cref{fig-robustness-to-corruption}, using Git-10M as an example, our method converges faster than the comparison methods across different pruning ratios. At different training steps, our method consistently achieves substantially lower FID than all competing approaches. With only about 50k training steps, our FID approaches the best scores of competitors at 100K.

\begin{figure}[t]
    \centering
    \includegraphics[width=0.99\linewidth]{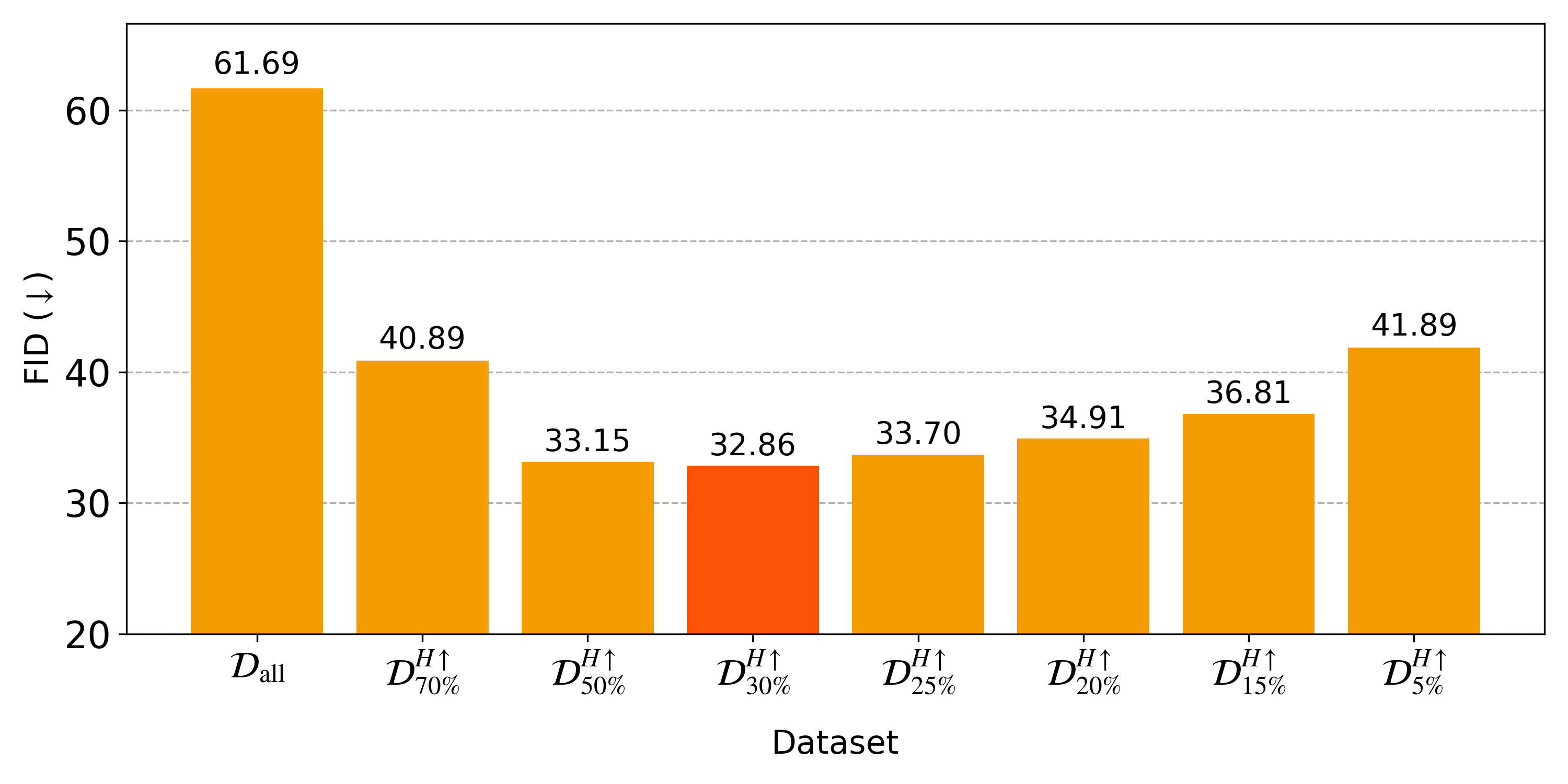}
    \caption{Results of entropy-based pruning at different pruning ratios on GiT-10M, reporting FID for the generation task.}
    \label{fig:entropy-based}
\end{figure}

\cref{tab:gen-metrics-two-datasets} presents results at a fixed pruning ratio of 70\% on both datasets. The proposed method achieves the best performance across generation and downstream tasks, demonstrating strong robustness and generalization. In addition, A$^{2}$\mbox{-}MAE, designed for global-scale datasets, exhibits a notable performance drop on the urban-scale USA-1m. On Git-10M, our method achieves the lowest generation FID (28.45), representing a 27.7\% improvement over the next best score (39.34). It further improves performance on super-resolution and semantic image synthesis, demonstrating strong alignment between downstream metrics and generation FID. On USA-1m, our approach also achieves consistent improvements across metrics, further confirming its effectiveness and generalizability to multispectral remote sensing imagery. More in-depth analyses regarding training steps are provided in supplementary materials.

\subsection{Ablation Study}

\subsubsection{Effectiveness of entropy-based pruning}

\begin{table}[t]
\centering
\caption{Results of stage I+II method at different pruning ratios.}
\label{tab:I+II}
\setlength{\tabcolsep}{2pt}      
\renewcommand{\arraystretch}{1.05} 
\scriptsize
\resizebox{\columnwidth}{!}{%
\begin{tabular}{@{}c|ccc|ccc@{}}
\toprule
\multirow{2}{*}{\makecell[c]{\\Pruning\\ratio}} &
\multicolumn{3}{c|}{\textbf{Git-10M}} &
\multicolumn{3}{c}{\textbf{RS5M}} \\
\cmidrule(lr){2-4}\cmidrule(lr){5-7}
&
\multicolumn{1}{c}{Generation} & \multicolumn{2}{c|}{SIS} &
\multicolumn{1}{c}{Generation} & \multicolumn{2}{c}{SIS} \\
&
FID$\downarrow$ & FID$\downarrow$ & LPIPS$\downarrow$ &
FID$\downarrow$ & FID$\downarrow$ & LPIPS$\downarrow$ \\
\midrule
0\%  & 127.8932 & 143.2622 & 0.6236 & 113.3864 & 137.9567 & 0.6076 \\
30\% & 100.3227 & 134.2329 & 0.6064 &  81.9246 & 122.9466 & 0.5950 \\
50\% &  84.3062 & 134.8372 & 0.6172 &  80.7609 & 121.7632 & 0.6002 \\
70\% &  70.0361 & 131.7129 & 0.6045 &  63.3343 & 125.0599 & 0.5950 \\
85\% &  \textbf{61.3269} & \textbf{120.5167} & \textbf{0.5982} &
       \textbf{58.9813} & \textbf{115.6430} & \textbf{0.5902} \\
90\% &  63.6400 & 123.0029 & 0.6106 &  64.2536 & 124.8206 & 0.6004 \\
\bottomrule
\end{tabular}
}
\end{table}

\begin{table*}[htbp]
\centering
\caption{Generative quality and runtime of clustering on full unlabeled data, entropy-pruned unlabeled data and our reference dataset-guided approaches.}
\label{tab:runtime-comparison-full}
\scriptsize 
\setlength{\tabcolsep}{5pt} 
\renewcommand{\arraystretch}{1}
\resizebox{0.8\textwidth}{!}{%
\begin{tabular}{c|c|c|c|c|c} 
\toprule
Method & \makecell{Cluster \\ sample size}  & \makecell{ Cluster\\Numbers} & \makecell{ Feature \\Dimension} &Times(s)& FID$\downarrow$ \\
\midrule
Full unlabeled clustering   & 10.5 million & 200 &1024& 4630.3  & 108.84 \\
\midrule
Entropy-pruned unlabeled clustering & 3.1 million & 200 &128& 308.4 &  71.16 \\
Reference-guided (single, 21 classes) & 2{,}100 & 200 & 1024  & 82.3 & 61.69 \\
Reference-guided (single, 45 classes) & 31{,}500 & 200  &1024& \textbf{76.6}  & \textbf{60.14} \\
Reference-guided (3 datasets)  & 43{,}600 & 200& 1024&101.8  & 62.28 \\
Reference-guided (5 datasets)  & 55{,}605 & 200 &1024& 115.1  & 60.65 \\
\bottomrule
\end{tabular}
}

\end{table*}

To validate the effectiveness of entropy-based pruning, we conduct experiments on the Git-10M dataset. Images are ranked in descending order of Shannon entropy, and the top $p\%$ subset is retained, denoted as $\mathcal{D}^{H\uparrow}_{p\%}$, while the unfiltered dataset is denoted as $\mathcal{D}_{\text{all}}$. Under a fixed training budget of 60K iterations, generation models trained on each subset are evaluated using FID, as shown in \cref{fig:entropy-based}. On this global-scale dataset, retaining only the top $30\%$ of images achieves the best FID (32.86), substantially outperforming the full dataset (61.69), highlighting the importance of removing low-information data. We further observe that as the pruning threshold becomes more stringent, FID initially decreases and then increases, indicating that the optimal pruning ratio should be chosen based on the redundancy level of the dataset. Analyses of different datasets and optimal pruning ratios are provided in supplementary materials.

\subsubsection{Effectiveness of scene-aware clustering and sampling}
In this section, we conduct experiments on two datasets with different scales, including GiT-10M (tens of millions) and RS5M (millions), to evaluate the effectiveness of stage II, as shown in \cref{tab:scene}. On GiT-10M, pruning at 85\% and 90\% leads stage II to outperform stage I alone by approximately 10\%. Similarly, on RS5M, stage II delivers notable benefits under high pruning, with corresponding improvements in downstream SIS performance. This is because, under high pruning ratios, stage I sampling alone leads to insufficient sample representativeness, and clustering is required to preserve diversity and distributional coverage. Overall, stage II consistently provides stable gains across both large- and medium-scale datasets, with especially pronounced advantages in highly compressed scenarios. And additional ablations on sampling strategies in stage II can be found in supplementary materials.

\cref{tab:I+II} further analyzes our two-stage method (stage~I + stage~II) under different pruning ratios. On both GiT-10M and RS5M, model performance steadily improves as the pruning ratio increases, with the best generative quality and downstream task results achieved at around 85\% pruning. When the pruning ratio reaches 90\%, the metrics exhibit a slight degradation, yet they remain clearly superior to training on the full unpruned dataset. These results indicate that our approach enables the training of strong pretrained models even after discarding the vast majority of training samples, and that its effectiveness is robust and transferable across both large- and medium-scale remote sensing datasets.
%


\subsubsection{Effectiveness of reference dataset-guided clustering method}

We further perform ablation studies on the proposed reference–guided clustering strategy in comparison with the standard practice of clustering directly on the unlabeled dataset. As reported in \cref{tab:runtime-comparison-full}, our approach consistently surpasses clustering the full unlabeled collection in both runtime efficiency and generative quality. Even the widely adopted MiniBatch KMeans variant, which reduces the feature dimensionality from 1024 to 128 with subsequent normalization, remains less effective. The improvement arises from exploiting expert-curated reference datasets to derive more representative scene prototypes, thereby facilitating faster sample assignment and enhancing FID.

We also examine the influence of reference dataset selection, varying in scene diversity and dataset scale, together with the associated degree of over-clustering. Experiments conducted with datasets of different complexity levels and their combinations yield several observations. An important finding is that the diversity of scene categories in the reference dataset plays a critical role, as datasets with broader coverage produce more representative clustering outcomes. Another key observation is that the number of clusters $K$ admits a moderate optimal range, with $K{=}200$ emerging as a robust configuration that adequately covers common RS scenes while avoiding the breakdown of scene-balanced sampling caused by excessive fragmentation. Finally, we note that a simple aggregation of multiple scene classification datasets is not automatically beneficial, since distributional discrepancies across datasets may impede the alignment of semantically related categories and thus undermine representativeness. Therefore, effective dataset integration requires consideration of both scene complementarity and domain compatibility. Due to the limited space, limitation and future work can be seen in supplementary materials.



\section{CONCLUSION}
In this work, we comprehensively explore data pruning for RS generative foundation models across datasets of varying scale and coverage. We propose a training-free, two-stage pruning strategy that jointly accounts for data heterogeneity, diversity, and representativeness. By avoiding the computational of large-scale clustering and model training, our approach enables rapid pruning of massive datasets into a high-quality subset under high pruning ratios, thereby accelerating the convergence of remote sensing generative foundation models and providing a strong data and modeling backbone for subsequent downstream applications. Our approach offers practical guidance and empirical findings that support the development of future remote sensing generative foundation models.
%

\newpage
{
    \small
    \bibliographystyle{ieeenat_fullname}
    \bibliography{main}
}


\end{document}